\documentclass{aamas2016}

\usepackage{amsmath}    % need for sub equations
\usepackage{amsfonts}
\usepackage{amssymb}
\usepackage{mathabx}
\usepackage{graphicx}   % need for figures
\usepackage{subcaption}
\usepackage{epsfig} 
\usepackage{color}
\usepackage{enumerate}
\usepackage{url}
\usepackage{cite}
\usepackage{booktabs}
\usepackage{todonotes}
\usepackage{ifthen,version}
\newboolean{include-notes}
\setboolean{include-notes}{true}
\DeclareRobustCommand{\adnote}[1]{\ifthenelse{\boolean{include-notes}}%
	{\textcolor{blue}{\textbf{AD: #1}}}{}}
\DeclareRobustCommand{\jbhnote}[1]{\ifthenelse{\boolean{include-notes}}%
	{\textcolor{red}{\textbf{JBH: #1}}}{}}
\DeclareRobustCommand{\JFFnote}[1]{\ifthenelse{\boolean{include-notes}}%
	{\textcolor[rgb]{0,0.7,0}{\textbf{JFF: #1}}}{}}
\DeclareRobustCommand{\clnote}[1]{\ifthenelse{\boolean{include-notes}}%
	{\textcolor[rgb]{1,0.5,0}{\textbf{CL: #1}}}{}}

\renewcommand{\H}[1]{\textbf{H#1}}
\newcommand{\anova}[4]{$F(#1, #2)\!=\!#3$, $#4$}
\newcommand{\ttest}[3]{$t(#1)\!=\!#2$, $#3$}
\newcommand{\chisquare}[3]{$\chi^2(#1)\!=\!#2$, $#3$}
\newcommand{\ztest}[2]{$z\!=\!#1$, $#2$}

\def\sharedaffiliation{%
\end{tabular}
\begin{tabular}{c}}

\begin{document}

\title{Goal Inference Improves Objective and Perceived Performance in Human-Robot Collaboration}

\numberofauthors{6}

\author{
\alignauthor
Chang Liu
\thanks{These authors contributed equally to this paper.\newline
This work is supported by ONR under the Embedded Humans MURI (N00014-13-1-0341).
}\ $\,^\S$\\
\email{changliu@berkeley.edu}
\alignauthor
Jessica B. Hamrick\ $^{*\dagger}$\\
\email{jhamrick@berkeley.edu}
\alignauthor Jaime F. Fisac\ $^{*\ddagger}$\\
\email{jfisac@berkeley.edu}
\and  % use '\and' if you need 'another row' of author names
\alignauthor Anca D. Dragan\ $^\ddagger$\\
\vspace{-2pt}\email{anca@berkeley.edu}
\alignauthor J. Karl Hedrick\ $^\S$\\
\email{khedrick@me.berkeley.edu}
\and
\alignauthor S. Shankar Sastry\ $^\ddagger$\\
\vspace{-2pt}\email{sastry@eecs.berkeley.edu}
\alignauthor Thomas L. Griffiths\ $^\dagger$\\
\email{tom\_griffiths@berkeley.edu}
\sharedaffiliation
\affaddr{$^\S$ Department of Mechanical Engineering}\\
\affaddr{$^\dagger$ Department of Psychology}\\
\affaddr{$^\ddagger$ Department of Electrical Engineering and Computer Sciences}\\
\affaddr{University of California, Berkeley}\\
\affaddr{Berkeley, California, 94720}
}

\maketitle

\begin{abstract}
The study of human-robot interaction is fundamental to the design and use of robotics in real-world applications.
Robots will need to predict and adapt to the actions of human collaborators in order to achieve good performance and improve safety and end-user adoption. 
This paper evaluates a human-robot collaboration scheme that combines the task allocation and motion levels of reasoning: the robotic agent uses Bayesian inference to predict the next goal of its human partner from his or her ongoing motion, and re-plans its own actions in real time. 
This anticipative adaptation is desirable in many practical scenarios, where humans are unable or unwilling to take on the cognitive overhead required to explicitly communicate their intent to the robot.
A behavioral experiment indicates that the combination of goal inference and dynamic task planning significantly improves both objective and perceived performance of the human-robot team. Participants were highly sensitive to the differences between robot behaviors, preferring to work with a robot that adapted to their actions over one that did not.

\end{abstract}

\keywords{Human-Agent Interaction, Teamwork, Intention Inference, Collaborative Task Allocation}

\begin{figure}[ht!]
	\begin{center}
		\includegraphics[width=0.5\textwidth,trim={6.5cm .7cm 4.6cm 0.7cm},clip] {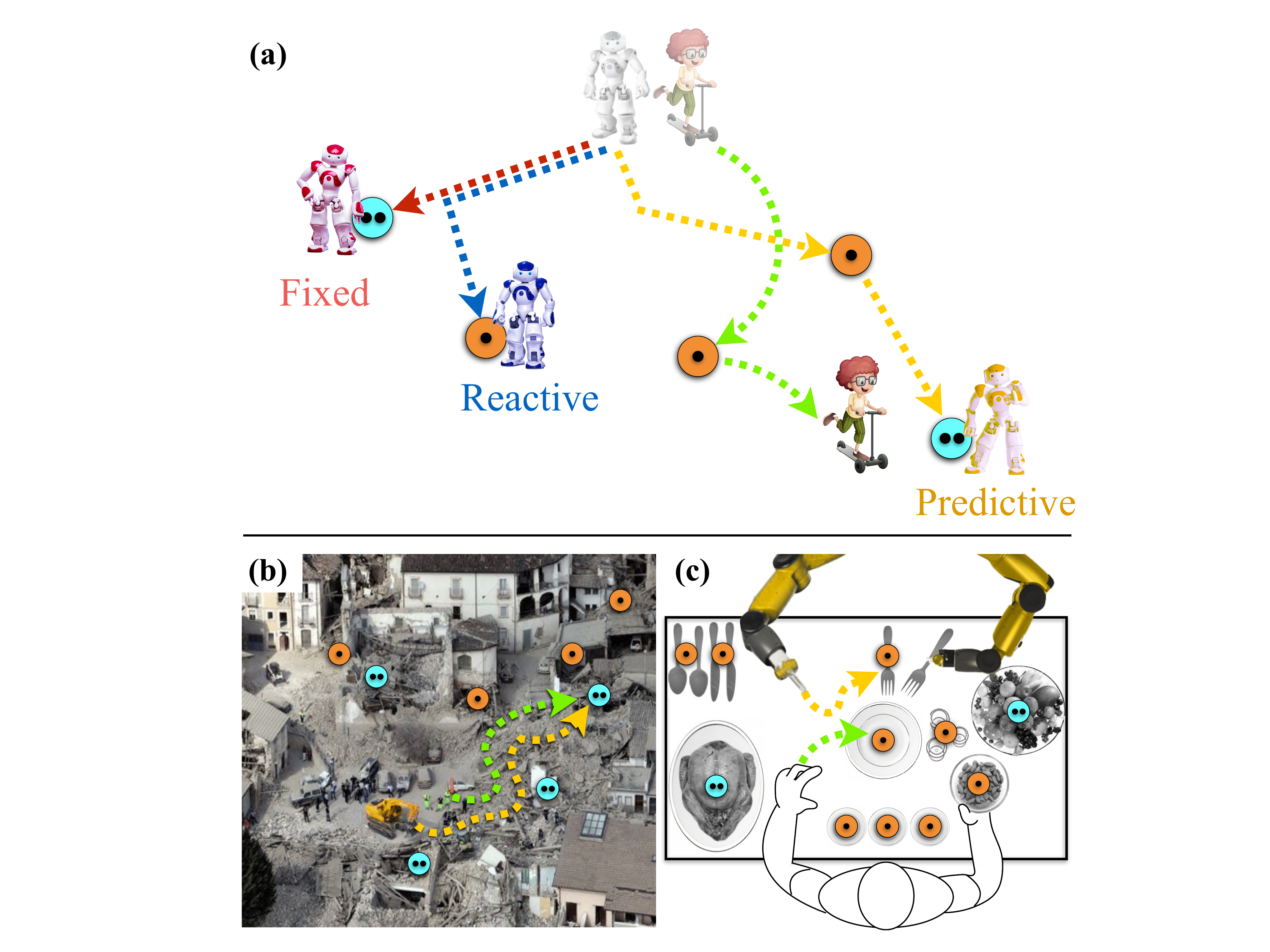}
		\caption{(a) Human avatar and three types of robot avatar used in the experiment. Circles with a single dot represent tasks that can be completed by any agent (either human or robot) and circles with two dots require simultaneous collaboration of both agents. (b)~Illustration of a search-and-rescue collaboration scenario. (c)~Illustration of a collaborative table  setting scenario.\vspace{-0.6cm}}
		\label{fig:scenario}
	\end{center}
\end{figure}

\section{Introduction}

Once confined to controlled environments devoid of human individuals, robots today are being pushed into our homes and workplaces by unfolding advances in autonomy.
From sharing workcells \cite{lenz2008joint}, to assisting older adults \cite{kidd2006sociable,wada2007living,bemelmans2012socially}, to helping in surveillance \cite{grocholsky2006cooperative} and search and rescue \cite{furukawa2006recursive,tisdale2009autonomous}, robots  need to operate in human environments, both with and around people.

While autonomous systems are becoming less reliant on human oversight, it is often still desirable for robots to collaborate with human partners in a peer-to-peer structure \cite{fong2005peer,dias2008sliding,shah2010empirical}.
In many scenarios, robots and humans can fill complementary roles in completing a shared task, possibly by completing different sub-goals in a coordinated fashion (Fig. \ref{fig:scenario}a).
For example, in a search-and-rescue mission (Fig. \ref{fig:scenario}b), human first responders might load injured persons onto an autonomous vehicle, which would then transport the victims to a medical unit.
In a number of manipulation scenarios, humans and robots can take advantage of their different aptitudes and capabilities in perception, planning and actuation %(speed, strength, precision, situational awareness)
to boost performance and achieve better results.

Collaboration hinges on coordination and the ability to infer each other's intentions and adapt \cite{tomasello2005understanding,vesper2010minimal}.
There is an abundant history of failure cases in human-automation systems due to coordination errors and miscommunication between human and autonomous agents \cite{McRuer1995,Sarter1997,Saffarian2012a}.
Such failures often result from an inability of autonomous systems to properly anticipate the needs and actions of peers; ensuring that autonomous systems have such an ability is critical to achieving good team performance \cite{yen2006agents}. 
Consider the collaborative manipulation scenario in Fig. \ref{fig:scenario}c, in which both the human and the robot work together to set a table.
It is important for the robot to anticipate the object that the human will take so that it can avoid hindering or even injuring the human.

Across robotics\cite{ziebart2009planning,dragan2012formalizing}, plan recognition\cite{charniak1993bayesian}, and cognitive science \cite{Baker2009,pezzulo2013human}, Bayesian goal inference has been applied to the problem of inferring human goals from their ongoing actions. 
Recently, we proposed a task allocation method which utilizes Bayesian goal inference at the motion level to enable the robot to adapt to the human at the task planning level \cite{Liu2014a}.
However, the strategy detailed in this work has not been evaluated in collaboration scenarios involving real humans.
Building on our initial work, we make the following contributions in this paper:

\noindent\textbf{Evaluation}: we evaluate the combination of two methods traditionally studied in isolation, namely applying goal inference at the motion level to compute a collaborative plan at the task level.
This work differs from other user studies evaluating the use of Bayesian goal inference in human-robot interaction, which have focused on assisting humans to achieve an immediate goal\cite{dragan2013policy,fern2014decision}.
We developed and ran a behavioral experiment (Fig. \ref{fig:scenario}a) in which participants collaborated with virtual robots that (i) performed goal inference and re-planned as the human was moving (the \textit{predictive} robot); (ii) only re-planned after the person reached their goal (the \textit{reactive} robot); or (iii) never re-planned unless the current plan became infeasible (the \textit{fixed} robot). We also had two between-subjects conditions in which we varied the type of inference used by the robot.  In the ``oracle'' condition, the predictive robot had direct access to the participant's intent, emulating perfect goal inference, while in the ``Bayesian'' condition, the predictive robot used probabilistic inference on noisy human trajectories (as would happen in a realistic setting). 
Building on literature that evaluates human-robot collaboration \cite{hoffman2007effects,dragan2015effects}, we evaluated performance both objectively and subjectively. Objective metrics included time to complete the task and the proportion of tasks completed by the robot; subjective metrics were based on elicited preferences for the robots, and a survey regarding perceptions of the collaboration.
As expected, errors in prediction do increase task completion time and lower subjective ratings of the robot. Despite these errors, both conditions produce the same results: we find that a robot that
anticipates human goals to adapt its overall task plan
is significantly superior to one that does not, in terms of both objective performance and subjective perception.

\noindent\textbf{Implication.}
Despite the nontrivial assumption that the human will plan optimally after the current goal, the behavioral study reveals that using this model for the robotic agent's adaptation greatly improves objective team performance (reducing task completion time and increasing the proportion of robot-completed goals). This may be due to the combination of swift adaptation to the human (which attenuates the impact of the optimality assumption) with an additional closed-loop synergy arising as the adapted robot plan becomes more likely to match the human's own expectation. Importantly, adaptation based on this model also significantly improves subjective perceptions: users patently notice---and appreciate---the agent's adaptive behavior. The implications of these results on human-robot interaction design are explored
in the discussion.

\section{Task Allocation During Human-Robot Collaboration}
\label{sec:formulation}

To test the importance of inferring human intent and adapting in collaborative tasks, we use a task allocation scenario with both individual and shared tasks.
Following \cite{Liu2014a}, we consider a setting in which human agent $H$ and a robot agent $R$ are in a compact planar domain $\mathcal{D}\subset\mathbb{R}^2$ with $N$ one-agent tasks located at positions $p_i\in\mathcal{D}$, $i=1,...,N$, and $M$ joint tasks with locations $q_j\in\mathcal{D}$, $j=1,...,M$.
One-agent tasks are executed instantly when visited by any agent, whereas joint tasks are only completed when visited by both agents simultaneously. 
The agent that arrives first at a joint task cannot leave until the other one comes to help complete the task. 
The human starts at position $h_0\in\mathcal{D}$ and the robot starts at $r_0\in\mathcal{D}$. 
Both agents are able to move in any direction at the same constant velocity $V$. 
The discrete-time dynamics of the human-robot team at each time step $k\ge 0$ are
$ h_{k+1} = h_k + u$ and $r_{k+1} = r_k + w$, respectively, where $\|u\|=\|w\|=V$, with $\|\cdot\|$ being the Euclidean norm.
Both agents have the common objective of completing all tasks in the minimum amount of time.
Each agent can observe its own location, the location of the other agent, and the location and type of all remaining tasks. 
However, agents can not explicitly communicate with each other.
This can be seen as a multi-agent version of the Traveling Salesman Problem \cite{held1970traveling}.

\subsection{Robot Collaboration Scheme}
\label{sec:adaptive-robot}

The scheme for integrating goal inference into task allocation presented in \cite{Liu2014a} achieves robustness to suboptimal human actions by adjusting its plan whenever the human's actions deviate from the robot's current internal allocation. 
At the initial time $k=0$, the robot computes the optimal task allocation and capture sequence given the initial layout of tasks and agent positions (Section \ref{sec:task-allocation}).
The robot observes the human's actions as both human and robot move towards their respective goals.
To infer whether the human is deviating from the robot's internal plan, the robot conducts a \textit{maximum a posteriori} (MAP) estimate (Section \ref{sec:goal-inference}).
If a deviation is detected, the robot computes a new allocation of tasks based on the current state and the inferred human intent.

When computing the optimal allocation, it is assumed that the human will in fact execute the resulting joint plan. In the ideal scenario where both agents are perfectly rational, this would indeed be the case (without the need for explicit coordination). Of course, this rationality assumption is far from trivial in the case of human individuals; we nonetheless make use of it as an approximate model, admitting incorrect predictions and providing a mechanism to adapt in these cases.
It is worth noting that this model maximally exploits collaboration in those cases in which the prediction is correct: that is, if the human does follow the computed plan, then the team achieves the best possible performance at the task. As will be seen, adaptation based on this model leads to significant improvements in team performance.

\subsection{Bayesian Goal Inference}
\label{sec:goal-inference}

Let $g$ denote the location of any of the $N+M$ tasks. 
Once the human starts moving towards a new task, the robot infers the human's intent as the task with the highest posterior probability, $P(g|h_{0:k+1})$, given the history of observations thus far, $h_{0:k+1}$.
The posterior probability is updated iteratively using Bayes' rule:
\begin{equation*}
P(g|h_{0:k+1})\ \propto\ P(h_{k+1}|g,h_{0:k}) P(g|h_{0:k}),
\end{equation*}
where the prior $P(g|h_{0:k})$ corresponds to the robot's belief that the human is heading towards task $g$ based on previous observed positions $h_{0:k}$.
The likelihood function $P(h_{k+1}|g,h_{0:k})$ corresponds to the transition probability for the position of the human given a particular goal.
Under the Markov assumption, the likelihood function can be simplified as: 
\begin{equation*}
P(h_{k+1}|g,h_{0:k}) = P(h_{k+1}|g,h_k).
\end{equation*}

We compute the term $P(h_{k+1}|g,h_k)$ assuming that the human is noisily optimizing some reward function. Following \cite{Baker2009}, we model this term as a Boltzmann (`soft-max') policy: 
\begin{equation*}
P(h_{k+1}|g,h_k)\ \propto\ \exp(\beta V_g(h_{k+1})),
\end{equation*}
where $V_g(h)$ is the value function for each task location $g$. Following \cite{Liu2014a}, we model $V_g(h)$ as a function of the distance between the human $h$ and the task, $d_g(h) = \|g-h\|$; i.e. $V_g(h) = \gamma^{d_g(h)} U - c[(\gamma-\gamma^{d_g(h)})/(1-\gamma)]$.
This corresponds to the value function of a $\gamma$-discounted optimal control problem with terminal reward $U$ and uniform running cost $c$.
The value function causes the Boltzmann policy to assign a higher probability to actions that take the human closer to the goal location $g$.

The parameter $\beta$ is a rationality index, capturing how diligent we expect the human to be in optimizing their reward.
When $\beta = 0$, $P(h_{k+1}|g,h_k)$ gives a uniform distribution, with all actions equally likely independent of their reward (modeling an indifferent human agent).
Conversely, as $\beta\to\infty$, the human agent is assumed to almost always pick the optimal action with respect to their current goal.

\subsection{Task Allocation}
\label{sec:task-allocation}

We compute the optimal allocation and execution order of tasks that minimizes completion time using a mixed-integer linear program (MILP).
Let $\mathbb{A}$ denote the set of human and robot agents and $\mathbb{G}$ represent the set of remaining tasks.
Then, the task allocation problem can be formulated as:
\begin{align*}
\min_{\mathbf{x}_{\rm alloc},\mathbf{t}_{\rm alloc}} \quad &\max_{v \in \mathbb{V}, g\in \mathbb{G}}  t^v_g \label{eqn:MILP}\\
\text{subject to} \quad &  \mathbf{x}_{\rm alloc} \in \mathbb{X}_{\rm feasible}\nonumber
\quad \mathbf{t}_{\rm alloc} \in \mathbb{T}_{\rm feasible},\nonumber
\end{align*}
where $t^v_g$ denotes the time when $v^\text{\textit{th}}$ agent completes task $g$; $\mathbf{x}_{\rm alloc}$ encodes the allocation of tasks to each agent and $\mathbf{t}_{\rm alloc}$ represents the time for agents to arrive and leave tasks; and $\mathbb{X}_{\rm feasible}$ and $\mathbb{T}_{\rm feasible}$ denote the corresponding sets of feasible values that satisfy the requirements on completing one-agent and joint tasks.
To generate an allocation that respects the intent of the human agent, the inferred goal is assigned as the first task for the human, which is enforced in $\mathbb{X}_{\rm feasible}$.
Other constraints of $\mathbb{X}_{\rm feasible}$ include 1) each task is assigned to the human or robot, 2) an agent can only visit and complete a task once 3) an agent can leave a task position at most once and can only do so after visiting it (continuity condition) and 4) no subtour exists for any agent.
The timing constraints of $\mathbb{T}_{\rm feasible}$ requires that a joint task be completed the time when both the human and the robot visit the task while a common task be executed immediately when an agent visits it.
Readers interested in details of the MILP formulation of this particular traveling salesman problem can refer to \cite{Liu2014a}. 
In this study, the MILP is solved by the commercial solver Gurobi \cite{gurobi}.

\subsection{Evaluation}

The method outlined above was evaluated in \cite{Liu2014a} through simulations in which human behavior was modeled as a mixture of optimality and randomness. 
Specifically, humans were modeled assuming that they would take the optimal action with probability $\alpha$, or take an action uniformly at random with probability $1-\alpha$. 
However, as true human behavior is much more sophisticated and nuanced than this, it is unclear how well these results would generalize to real humans. 
Moreover, the simulation results tell us little about what people's perceived experiences might be.
Even if Bayesian goal inference does decrease the time needed to complete a set of tasks with real people, it is important that those people subjectively perceive the robot as being helpful and the collaboration as being fluent, and that they feel confident that they can rely on the robot's help and comfortable collaborating with it.

To more rigorously test the robot's task allocation strategy, we ran a behavioral experiment in which we asked human participants to collaborate with three different robots, as shown in Fig. \ref{fig:scenario}a, one at a time and in a counter-balanced order: a robot that did not re-plan its strategy regardless of the human's choices (the \textit{fixed} robot), a robot that reacted to these choices only once the human reached the goal in question (the \textit{reactive} robot), and a robot that anticipated the human's next goal at each step and adapted its behavior early (the \textit{predictive} robot).
We conducted the experiment under two different conditions:
one with an \textit{oracle-predictive} robot that can make perfect inferences of the human's intent, and a second with a \textit{Bayesian-predictive} robot, in which the Bayesian intent inference is at times erroneous (as it will be in real human-robot interactions).

\section{Experimental Design}
\label{sec:exp1}

\subsection{Task}

We designed a web-based human-robot collaboration experiment where the human participant was in control of an avatar on a screen.
In this experiment, the human and virtual robot had to complete a set of tasks as quickly as possible.
While some of the tasks could be completed by either the human or the robot (\textit{one-agent tasks}), others required both agents to complete the task simultaneously (\textit{joint tasks}).
The participant chose to complete tasks by clicking on them, causing the human avatar to move to that task while the robot moved towards a task of its choice.
After the human completed the chosen task, they chose another task to complete, and so on until all tasks were completed.

\subsection{Independent Variables}

To evaluate the collaboration scheme described in Section~\ref{sec:formulation}, we manipulated two  variables: \textbf{adaptivity} and \textbf{inference type}.

\subsubsection{Adaptivity}
We manipulated the amount of information that the robot receives during the task, which directly affects how adaptive the robot is to human actions.
The following robots vary from least adaptive to most adaptive:

\noindent\textbf{Fixed}: \textit{This robot receives no information about the human's goal.}
It runs an initial optimal task allocation for the human-robot team and subsequently executes the robot part of the plan irrespective of the human's actions.
It only re-plans its task allocation if an inconsistency or deadlock is reached.
For example, if the human finishes a task that the robot had internally allocated to itself, the fixed robot removes the task from its list of pending tasks and chooses the next pending task as the goal.
Additionally, if both agents reach different joint tasks, the robot moves to the human's location and re-plans its strategy to account for this deviation.

\noindent\textbf{Reactive}: \textit{This robot receives information about the human's goal only when the human reaches their goal.}
At this point, the robot reruns task allocation for the team assuming that the human will behave optimally for the remainder of the trial.

\noindent\textbf{Predictive}: \textit{This robot receives information about the human's goal before they reach the goal.}
After receiving this information, the robot recomputes the optimal plan, again assuming optimal human behavior thereafter.

\subsubsection{Inference Type}
We also manipulated the type of inference used by the predictive robot:

\noindent\textbf{Oracle-predictive}: \textit{This robot receives perfect knowledge of the human's next goal as soon as the human chooses it.}
Although there are real-world cases in which a human might inform a robot about its goal immediately after deciding, it is rare for a robot to know exactly the human's goal without having the human explicitly communicate it.

\noindent\textbf{Bayesian-predictive}: \textit{This robot continuously receives information related to the human's goal by observing the motion of the human.}
Based on this motion, the robot attempts to predict the human's goal using the Bayesian inference scheme described in Section \ref{sec:goal-inference}.
In most cases of human-robot collaboration, the robot will only have access to the human's motions.

\subsection{Experimental Conditions}

\begin{table}
	\centering
	\caption{Experimental Conditions}
	\label{tbl:conditions}
	\begin{tabular}[h]{ccc}
		\toprule
		\textbf{Inference Type} & \textbf{Adaptivity} & \textbf{Robot}\\
		(between-subjects) & (within-subjects) & \\
		\midrule
		- & Low Adaptivity & \textit{Fixed} \\
		- & Med. Adaptivity & \textit{Reactive} \\
		Oracle & High Adaptivity & \textit{Oracle-predictive} \\
		\midrule
		- & Low Adaptivity & \textit{Fixed} \\
		- & Med. Adaptivity & \textit{Reactive} \\
		Bayesian & High Adaptivity & \textit{Bayesian-predictive} \\
		\bottomrule
	\end{tabular}
\end{table}

We manipulated the independent variables of adaptivity and inference type in a $3\times 2$ mixed design in which inference type was varied between-subjects, while adaptivity was varied within-subjects.
All participants collaborated with the fixed and reactive robots (corresponding to the \textit{low adaptivity} and \textit{medium adaptivity} conditions, respectively).
For the \textit{high adaptivity} conditions, half the participants were assigned to a \textit{oracle} condition, in which they collaborated with the oracle-predictive robot; and half were assigned to a \textit{Bayesian} condition, in which they collaborated with the Bayesian-predictive robot.
This design is given in Table~\ref{tbl:conditions} for reference. 

The motivation for this design was to prevent participants from having to keep track of too many robots (as would be the case in a $4\times 1$ design), while still creating the experience of comparing the predictive robot with non-predictive baselines.

\subsection{Procedure}
\label{sec:exp1-procedure}

\subsubsection{Stimuli}
\label{sec:stimuli}
There were 15 unique task layouts which consisted of a collection of five or six tasks, as well as initial positions of the human and robot avatars. 
As shown in Fig. \ref{fig:scenario}a, one-agent tasks were depicted by an orange circle with a single dot inside it, and joint tasks were depicted by a cyan circle with two dots inside it.
The human avatar was a gender-neutral cartoon of a person on a scooter, and the robot avatars were images of the same robot in different poses and colors (either red, blue, or yellow), which were counterbalanced across participants.

\subsubsection{Trial Design}
\label{sec:trial-design}
First, we presented participants with four example trials to help them understand the task, which they completed with the help of a gray robot.
Each experimental layout was presented to participants three times over the course of the experiment (once with each of the fixed, reactive, and predictive robots).
The trials were grouped into five blocks of nine trials (45 trials total).
Within each block, participants completed three trials with one of the robots, then three with a different robot, and then three with the final robot.
The order of the robots and task layouts were randomized according to a Latin-square design, such that the order of the robots varied and such that no task layout was repeated within a block.

\subsubsection{Layout Generation}
\label{sec:layout-generation}
We randomly generated task layouts according to the following procedure.
First, we assumed a model of human goal selection that sets the probability of choosing the next goal task as inversely proportional to the distance between the position of the human agent and the tasks\footnote{Specifically, we used a Boltzmann policy to model human task selections. We fit the tuning parameter from participant's choices in an earlier pilot experiment using maximum-likelihood estimation, resulting in a best fitting parameter value of $\beta=1.05$.}.
Under this model, we calculated the expected task completion time of 104 randomly generated layouts with reactive and Bayesian-predictive robots.
We then computed the rank ordering of these completion times relative to all possible completion times for each layout, and computed the ratio of the rank order of the Bayesian-adaptive robot to the rank order of the reactive robot.
This ratio gave us a measure of how differently we would expect the two robots to behave: layouts with ratio greater than 1.5 (9 layouts) or less than 0.6 (6 layouts) were selected, giving a total of 15 layouts for each robot. Subjects completed 5 blocks of 9 layouts, totaling 45 layouts.

\subsubsection{Avatar Animation}
\label{sec:animation}
When moving to complete a task, the robot avatar always moved in a straight line towards the task.
For the human avatar, however, the trajectory followed was not a straight path, but a B\'ezier curve that initially deviated to a side and then gradually corrected its course towards the correct task location.
The reason for using this trajectory was to allow the Bayesian-predictive robot to make errors: if the human always moved in a straight line, then the Bayesian-predictive robot would almost always infer the correct goal right away (and thus would give nearly identical results as the oracle-predictive robot).
Furthermore, this choice of curved trajectory is a more realistic depiction of how actual humans move, which is not usually in a straight line \cite{Brogan2003}.

\subsubsection{Attention Checks}
\label{sec:attention}
After reading the instructions, participants were given an attention check in the form of two questions asking them the color of the one-agent tasks and joint tasks. 
At the end of the experiment, we also asked them whether they had been paying attention to the difference in helpfulness between the three robots.

\subsubsection{Controlling for Confounds}
We controlled for confounding factors by counterbalancing the colors of the robots (Section~\ref{sec:stimuli}); by using a different color robot for the example trials (Section~\ref{sec:trial-design}); by randomizing the trial order (Section~\ref{sec:trial-design}); by generating stimuli to give advantages to different robots (Section~\ref{sec:layout-generation}); by animating the human avatar in a realistic way to differentiate between the Bayesian-predictive and oracle-predictive robots (Section~\ref{sec:animation}); and by including attention checks (Section~\ref{sec:attention}).

\subsection{Dependent Measures}

\subsubsection{Objective measures}
\label{sec:objective-measures}
During each trial, a timer on the right side of the screen kept track of the time it took to complete the tasks. 
This timer was paused whenever the participant was prompted to choose the next task and only counted up when either the human avatar or robot avatar were moving.
The final time on the timer was used an object measure of how long it took participants to complete each trial.
We also tracked how many one-agent tasks were completed by the robot, the human, or both simultaneously.

\subsubsection{Subjective measures} 
\label{sec:subjective-measures}
After every nine trials, we asked participants to choose the robot that they most preferred working with based on their experience so far.
At the end of the experiment, we also asked participants to fill out a survey regarding their subjective experience working with each of the robot partners (Table~\ref{tbl:survey}).
These survey questions were based on Hoffman's metrics for fluency in human-robot collaborations \cite{Hoffman2013}, and were answered on a 7-point Likert scale from 1 (``strongly disagree") to 7 (``strongly agree").
We also included two forced-choice questions, which had choices corresponding to the three robots.

\subsection{Hypotheses}

\begin{table}
	\caption{Subjective Measures}
	\label{tbl:survey}
	\begin{tabular}[t]{p {.46\textwidth}}
		\toprule
		\textbf{Fluency}\\
		1. The human-robot team with the [color] robot worked fluently together.\\
		2. The [color] robot contributed to the fluency of the team interaction.\\
		\midrule
		\textbf{Contribution}\\
		1. (Reversed) I had to carry the weight to make the human-robot team with the [color] robot better.\\
		2. The [color] robot contributed equally to the team performance.\\
		3. The performance of the [color] robot was an important contribution to the success of the team.\\
		\midrule
		\textbf{Trust}\\
		1. I trusted the [color] robot to do the right thing at the right time.\\
		2. The [color] robot is trustworthy.\\
		\midrule
		\textbf{Capability}\\
		1. I am confident in the ability of the [color] robot to help me.\\
		2. The [color] robot is intelligent.\\
		\midrule
		\textbf{Perceptiveness (originally `Goals')}\\
		1. The [color] robot perceived accurately what my goals were.\\
		2. (Reversed) The [color] robot did not understand what I was trying to accomplish.\\
		\midrule
		\textbf{Forced-Choice Questions}\\
		1. If you were to redo this experiment with only one of the robots, which would you most prefer to work with?\\
		2. If you were to redo this experiment with only one of the robots, which would you least prefer to work with?\\
		\bottomrule
	\end{tabular}
\end{table}

We hypothesized that adaptivity and inference type would affect both objective measures (Section~\ref{sec:objective-measures}) and subjective measures (Section~\ref{sec:subjective-measures}).

\noindent\textbf{\H1 - Task Completion Time} \textit{Robot adaptivity will negatively affect task completion times, with the least adaptive robots resulting in the slowest completion times, and the most adaptive robots resulting in the fastest completion times. The Bayesian-predictive will robot result in slower completion times than the oracle-predictive robot.}

\noindent\textbf{\H2 - Robot Task Completion} \textit{Robot adaptivity will positively affect the number of tasks that the robot completes, with more adaptive robots completing more tasks than the less adaptive robots. The Bayesian-predictive robot will complete fewer tasks than the oracle-predictive robot.}

\noindent\textbf{\H3 - Robot Preference} \textit{Robot adaptivity will positively affect participants' preferences, with more adaptive robots being preferable to less adaptive robots. The Bayesian-predictive robot will be less preferable than the oracle-predictive robot. Additionally, preferences will become stronger as participants gain more experience with each robot.}

\noindent\textbf{\H4 - Perceptions of the Collaboration} \textit{Robot adaptivity will positively affect participants' perceptions of the collaboration as indicated by the subjective survey measures, with more adaptive robots being rated higher than less adaptive robots. Inference type will also affect perceptions, with the Bayesian-predictive robot scoring lower than the oracle-predictive robot.}

\subsection{Participants}

We recruited a total of 234 participants from Amazon's Mechanical Turk using the psiTurk experimental framework \cite{Gureckis15}.
We excluded 32 participants from analysis for failing the attention checks, as well as 6 due to an experimental error, leaving a total of $N=196$ participants whose data we used.
All participants were treated in accordance with local IRB standards and were paid \$1.20 for an average of 14.7 minutes of work, plus an average bonus of \$0.51. Bonuses could range from \$0.00 to \$1.35 depending on performance: for each trial, participants could get a \$0.03 bonus if they completed all tasks in the shortest possible amount of time; a \$0.02 bonus if they finished within the top 5\% fastest times; or a \$0.01 bonus if they finished within the top 10\% fastest times\footnote{When comparing completion times to the range of possible times, we always compared the participants' completion times with the range of all times achievable with the robot they were currently collaborating with.}. They received no bonus if they were slower than the top 10\% fastest times.

\section{Results}

\subsection{Manipulation Checks}

The goal of changing the inference type was ultimately to influence the number of errors made by the robot.
To ensure that the Bayesian-predictive robot did make more errors than the oracle-predictive robot, we counted the proportion of timesteps on each trial during which the Bayesian-predictive robot had an incorrect belief about the human's goal.
We then constructed a mixed-effects model for these error rates with only the intercept as fixed effects, and both participants and stimuli as random effects.
The intercept was $0.27\pm 0.021\ \mathrm{SE}$ (\ttest{14.043}{12.83}{p<0.001}), indicating an average error rate of about 27\%.
We also looked at the stimuli individually by reconstructing the same model, except with stimuli as fixed effects.
Using this model, we found that error rates ranged from $15.0\%\pm 1.18\%\ \mathrm{SE}$ to $41.0\% \pm 1.18\%\ \mathrm{SE}$.

\subsection{\H1 - Task Completion Time}

\begin{figure}[t]
	\centering
	\includegraphics[width=0.48\textwidth]{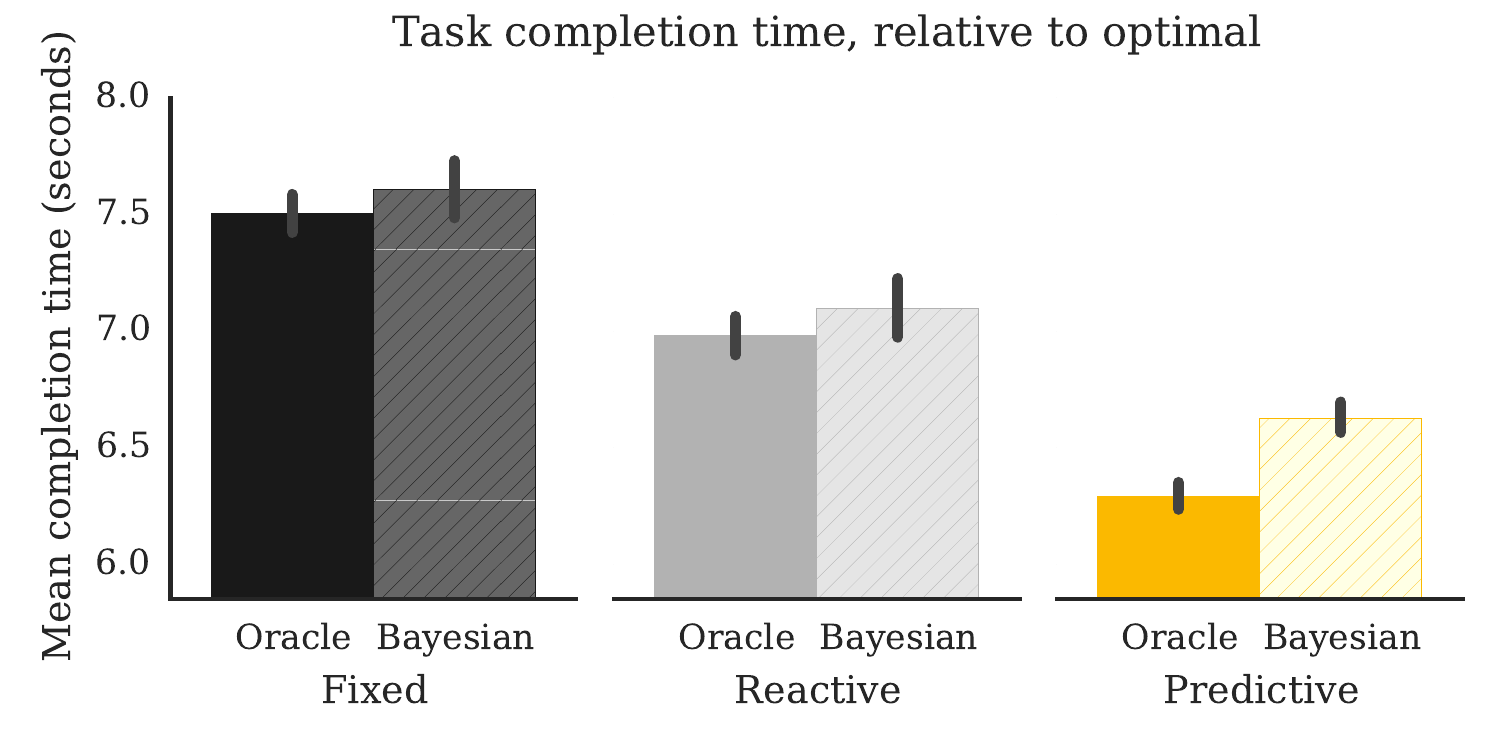}
	\caption{Comparison of completion times for different robots. The $y$-axis begins at the average optimal completion time, and thus represents a lower bound. Error bars are bootstrapped 95\% confidence intervals of the mean using 1,000 samples.\vspace{-0.5cm}}
	\label{fig:completion-time}
\end{figure}

First, we looked at the objective measure of task completion time, which is also shown in Fig.~\ref{fig:completion-time}.
To analyze task completion time, we first found the average completion time for each participant across stimuli, then constructed a linear mixed-effects model for these times using the inference type and adaptivity as fixed effects and the participants as random effects.
We found main effects of both inference type (\anova{2}{194}{10.30}{p<0.01}) and adaptivity (\anova{2}{388}{709.27}{p<0.001}), as well as an interaction between the two (\anova{2}{388}{10.07}{p<0.001}).

We performed post-hoc comparisons using the multivariate $t$ method for $p$-value adjustment.
In support of \H1, we found that the reactive robot led to completion times that were $0.52s\pm 0.029\ \mathrm{SE}$ faster than the fixed robot (\ttest{388}{17.777}{p\!<\!0.001}).
Also in support of \H1, the oracle-predictive robot led to completion times that were $0.69s\pm 0.041\ \mathrm{SE}$ faster than the reactive robot (\ttest{388}{-16.785}{p\!<\!0.001}), and the Bayesian-predictive robot led to completion times that were $0.47s\pm 0.041\ \mathrm{SE}$ faster than the reactive robot (\ttest{388}{-11.339}{p\!<\!0.001}).
There was no evidence for a difference between the fixed robots between participants (\ttest{329.53}{-1.528}{p\!=\!0.61}), nor between the reactive robots (\ttest{329.53}{-1.756}{p\!=\!0.46}).
There was, however, a difference between the oracle-predictive and Bayesian-predictive robots, (\ttest{329.53}{-5.03}{p\!<\!0.001}), with the oracle-predictive robot resulting in completion times that were faster by $0.34s\pm 0.067\ \mathrm{SE}$.
The combination of these results fully support \H1: when adaptivity increases or the number of errors in inference decreases, task completion time decreases.

\subsection{\H2 - Robot Task Completion}

\begin{figure}[t]
	\centering
	\includegraphics[width=0.48\textwidth]{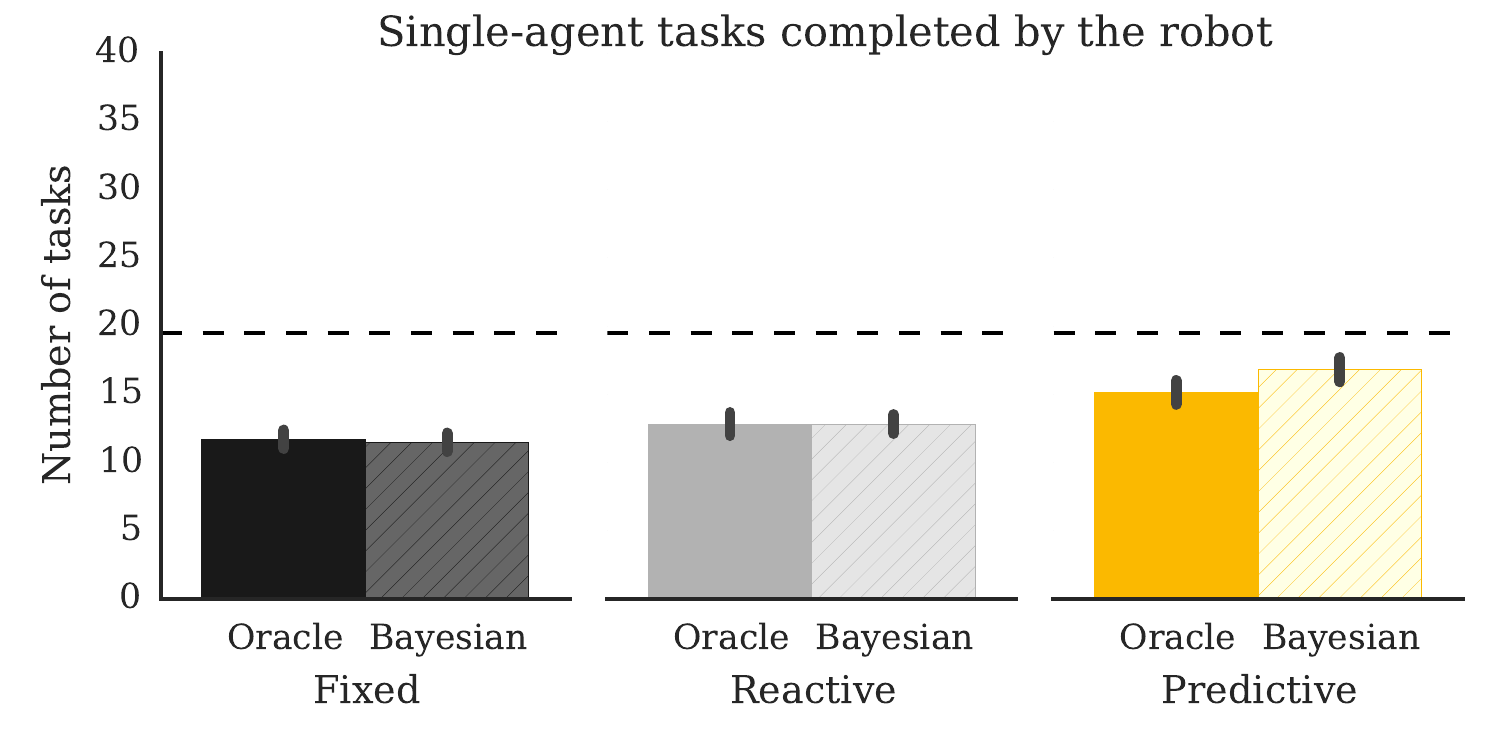}
	\caption{Number of one-agent tasks completed by the robot. The dotted lines represent the robot and the human completing an equal proportion of tasks. Error bars are bootstrapped 95\% confidence intervals of the mean using 1,000 samples.\vspace{-0.5cm}}
	\label{fig:tasks-completed}
\end{figure}

Next, we looked at the additional object measure of how many one-agent tasks were completed by the robot.\footnote{Sometimes the robot and human could complete the same one-agent task simultaneously. We excluded these cases from the number of one-agent tasks completed by the robot, because for these tasks the robot was not actually providing any additional help. We found that the robot and human competed for the same task 1.5\%, 8.6\%, 14.6\%, and 18.1\% of the time in the oracle-predictive, Bayesian-predictive, reactive, and fixed conditions, respectively.}
For each participant, we counted the total number of one-agent tasks completed by the robot throughout the entire experiment.
These numbers are shown in Fig.~\ref{fig:tasks-completed}.
To analyze these results, we constructed a linear mixed-effects model for the number of tasks with inference type and adaptivity as fixed effects and the participants as random effects.
There was no evidence for a main effect of inference type on the number of tasks completed by the robot (\anova{2}{194}{0.905}{p\!=\!0.34}), but there was an effect of the adaptivity on the number of tasks completed by the robot (\anova{2}{388}{171.15}{p\!<\!0.001}), as well as an interaction between adaptivity and inference type (\anova{2}{388}{9.237}{p\!<\!0.001}).

To investigate these effects further, we performed post-hoc comparisons with multivariate $t$ adjustments.
Consistent with \H2, we found that the reactive robot completed $1.2\pm 0.24\ \mathrm{SE}$ more tasks than the fixed robot (\ttest{388}{4.969}{p\!<\!0.001});
the oracle-predictive robot completed $2.3\pm 0.34\ \mathrm{SE}$ more tasks than the reactive robot  (\ttest{388}{6.722}{p\!<\!0.001}); and the Bayesian-predictive robot completed $4.0\pm 0.35\ \mathrm{SE}$ more tasks than the reactive robot (\ttest{388}{11.566}{p\!<\!0.001}).
Surprisingly, however, in comparing between subjects, we found the opposite of \H2: the Bayesian-predictive robot completed $1.7\pm 0.58\ \mathrm{SE}$ more tasks than the oracle-predictive robot (\ttest{316.19}{2.903}{p\!<\!0.05}).

These results confirm part of \H2---that adaptivity increases the number of tasks completed by the robot---but disconfirm the other part of \H2---that Bayesian inference decreases the number of tasks completed by the robot.
However, we note that the effects are quite small between the different robots:
the differences reported are in the order of 1--4 tasks 
for the entire experiment, which had a total of 39 one-agent tasks.
Each trial only had two or three one-agent tasks, and thus these results may not generalize to scenarios that include many more tasks.
Furthermore, although we found that the Bayesian inference type increased the number of tasks completed by the robot, we speculate that this may be due to the Bayesian-predictive robot inferring the incorrect human goal and therefore taking some other one-agent task. However, further investigation is required to determine the actual source of this difference.

\subsection{\H3 - Robot Preference}

We next analyzed the subjective measure of participants' preferences for which robot they would rather work with.

\subsubsection{Preferences over time}
\begin{figure}[t]
	\centering
	\includegraphics[width=0.48\textwidth]{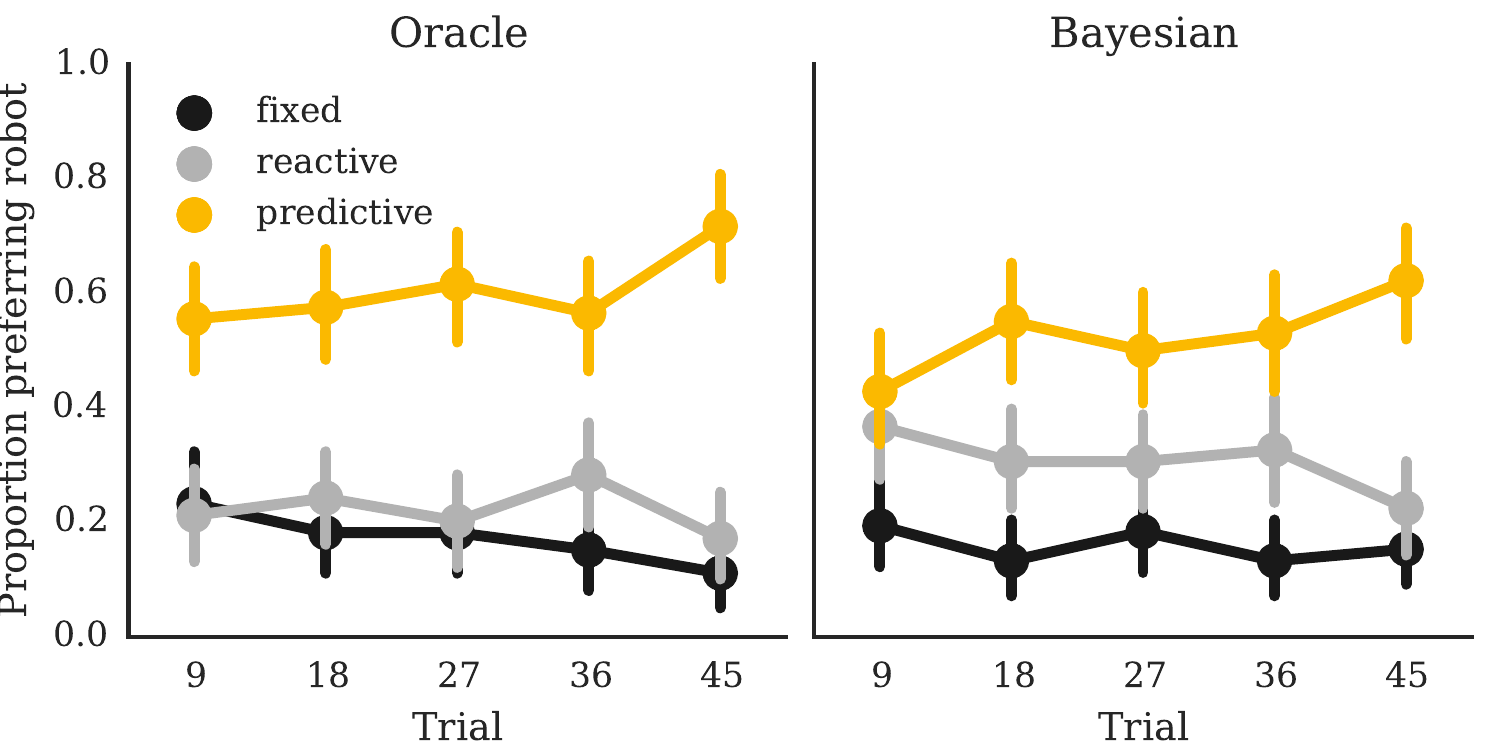}
	\caption{Preferred robot as a function of trial. Error bars are bootstrapped 95\% confidence intervals of the mean using 1,000 samples.\vspace{-0.5cm}}
	\label{fig:preferences}
\end{figure}

Fig.~\ref{fig:preferences} shows the proportion of participants choosing each robot as a function of trial.
We constructed a logistic mixed-effects model for binary preferences (where 1 meant the robot was chosen, and 0 meant it was not) with inference type, adaptivity, and trial as fixed effects and participants as random effects. Using Wald's tests, we found a significant main effect of adaptivity (\chisquare{2}{16.7184}{p\!<\!0.001}) and trial (\chisquare{1}{5.1104}{p\!<\!0.05}), supporting \H3.
We also found an interaction between adaptivity and trial (\chisquare{1}{9.2493}{p\!<\!0.01}).
While the coefficient for trial was negative ($\beta=-0.022\pm 0.0095\ \mathrm{SE}$, \ztest{-2.261}{p\!<\!0.05}), the coefficient for the interaction of trial and the predictive robot was positive ($\beta=0.036\pm 0.0120$, \ztest{3.017}{p\!<\!0.01}).
This also supports \H3: if preferences for the predictive robot increase over time, by definition the preferences for the other robots must decrease, thus resulting in an overall negative coefficient.

Although we did not find a significant main effect of inference type (\chisquare{1}{1.457}{p\!=\!0.23}), we did find an interaction between inference type and adaptivity (\chisquare{2}{7.7208}{p\!<\!0.05}).
Post-hoc comparisons with the multivariate $t$ adjustment indicated that there was a significant overall improvement in preference between the fixed and reactive robots (\ztest{-5.23}{p<0.001}) as well as an improvement in preference between the reactive and oracle-predictive robots (\ztest{11.86}{p\!<\!0.001}) and between the reactive and Bayesian-predictive robots (\ztest{6.95}{p\!<\!0.001}).
While there was no significant difference between participants for the fixed robot (\ztest{0.40}{p\!=\!0.998}), there was a significant difference for the reactive robot (\ztest{-2.93}{p\!<\!0.05}) and a marginal difference between the predictive robots (\ztest{2.52}{p\!=\!0.098}), with the oracle-predictive robot being slightly more preferable to the Bayesian-predictive robot.

\subsubsection{Final rankings}
For each participant, we assigned each robot a score based on their final rankings. The best robot received a score of 1; the worst robot received a score of 2; and the remaining robot received a score of 1.5.
We constructed a logistic mixed-effects model for these scores, with inference type and adaptivity as fixed effects and participants as random effects; we then used Wald's tests to check for effects.
In partial support of \H3, there was a significant effect of adaptivity on the rankings (\chisquare{2}{52.352}{p\!<\!0.001}) but neither a significant effect of inference type (\chisquare{1}{1.149}{p\!=\!0.28}) nor an interaction between adaptivity and inference type (\chisquare{2}{3.582}{p\!=\!0.17}).
Post-hoc comparisons between the robot types were adjusted with Tukey's HSD and indicated a preference for the reactive robot over the fixed robot (\ztest{5.092}{p\!<\!0.001}) and a preference for the predictive robots over the reactive robot (\ztest{5.634}{p\!<\!0.001}).

\subsection{\H4 - Perceptions of the Collaboration}

\begin{figure*}
	\centering
	\includegraphics[width=1\textwidth]{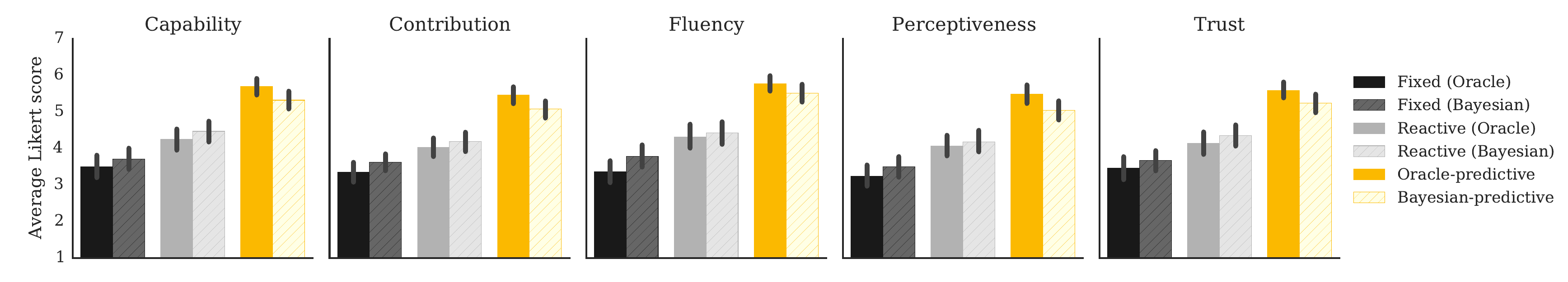}
	\caption{Findings for subjective measures. Error bars are bootstrapped 95\% confidence intervals of the mean.\vspace{-0.5cm}}
	\label{fig:rating_av}
\end{figure*}

Finally, we looked at the subjective measures of participants' perceptions of the collaboration.
For each robot and measure (capability, contribution, fluency, goals, and trust), we averaged each participants' responses to the individual questions, resulting in a single score per participant, per measure, per robot.
The results are shown in Fig.~\ref{fig:rating_av}.
We constructed a linear mixed-effects model for the survey responses with the adaptivity, inference type, and measure type as fixed effects, and with participants as random effects.
We found significant main effects for both the adaptivity (\anova{2}{2730}{543.18}{p\!<\!0.001}) and measure type (\anova{4}{2730}{5.28}{p\!<\!0.001}).
While there was no significant main effect of inference type (\anova{2}{195}{0.08}{p\!=\!0.78}), there was an interaction between adaptivity and inference type (\anova{2}{2730}{17.17}{p\!<\!0.001}).
We did not find interactions between inference type and measure type (\anova{4}{2730}{0.14}{p\!=\!0.97}), between adaptivity and measure type (\anova{8}{2730}{0.38}{p\!=\!0.93}), or between adaptivity, inference type, and measure type (\anova{8}{2730}{0.14}{p\!=\!0.997}).

We investigated these effects further through a post-hoc analysis with multivariate $t$ corrections.
In support of \H4, we found that participants rated the reactive robot higher than the fixed robot by $0.72\pm 0.058\ \mathrm{SE}$ points on the Likert scale (\ttest{2730}{12.415}{p\!<\!0.001}).
They also rated the oracle-predictive robot higher than the reactive robot by $1.44\pm 0.082\ \mathrm{SE}$ points (\ttest{2730}{17.568}{p\!<\!0.001}), the Bayesian-predictive robot higher than the reactive robot by $0.91\pm 0.083\ \mathrm{SE}$ points (\ttest{2730}{11.064}{p\!<\!0.001}), and the oracle-predictive robot higher than the Bayesian-predictive robot by $0.36\pm 0.112$ points (\ttest{470.31}{3.267}{p\!<\!0.05}).

\section{Discussion}

In this paper we set out to answer three questions. First, does an increase in robot adaptivity lead to an objective improvement in real human-robot collaboration scenarios?
Second, even if performance improves objectively, do people actually notice a difference?
Third, if the robot does not have perfect knowledge of the human's goals and must infer them, does adaptivity still result in the same improvements?

\textbf{Implications.} Our results suggest that outfitting robots with the ability to adaptively re-plan their actions from a motion-based inference of human intent can lead to a systematic improvement in both the objective and perceived performance in human-robot collaboration.
We found that this type of adaptivity increases performance objectively by lowering the amount of time it takes the team to complete the tasks and by improving the balance of work between the two agents.
Subjectively, we found that participants preferred motion-based predictive adaptation to purely reactive adaptation at the task level (which, as expected, was in turn preferred to no adaptation), and that they rated the more adaptive robot higher on a number of measures including capability, collaboration, fluency, perceptiveness, and trust.
Unsurprisingly, introducing prediction errors through manipulation of the inference type generally decreased both objective performance and subjective ratings.
What is surprising, however, is how small this decrease in performance was, particularly given that the Bayesian-predictive robot had an error rate ranging from a low of 15\% to a high of 40\% for some stimuli.
This suggests that an attempt to predict and adapt to human goals---even under limited accuracy---can have a significant effect on the team's objective performance and the human's perception of the robot.

Another non-obvious observation is that the adaptation mechanism is successful in spite of 
the strong assumption that the human agent will behave optimally from the current state. In practice, the impact of this assumption is limited by the fact that the robot quickly adapts whenever the human does not follow this optimal plan; this is comparable to how model-predictive control can use a nominal system model and correct for model mismatch at each re-planning step. Additional effects may also explain the observed improvement. In particular, our model of human task planning explicitly accounts for the joint actions of the two agents (as opposed to only their own); because human inferences of other agent's intentions have been shown to be consistent with models of approximately rational planning \cite{Baker2009}, it is therefore likely that the robot's adapted actions closely match users' actual expectations.
In addition, the fact that the robot adapts by switching to the new joint optimal plan means that the human is always ``given a new chance'' to behave optimally; in other words, the robot's adaptation is always placing the team in the situation with the best achievable outcome. Developing and testing these hypotheses further will be the subject of future work.

\textbf{Limitations.} There are several limitations to this work which we plan to address in future research. First, in our experiment, participants could only select the goal task and the trajectory for human avatar was predefined. It would be more realistic if participants were allowed to directly control the motion of human avatar.
To better understand the role of realistic control, we will conduct a future study in which participants can continuously move their avatar using the arrow keys.
Second, the robot's working assumption that the human will follow the optimal joint plan is unrealistic. Better results could conceivably be obtained if this assumption were to be relaxed and replaced by a more refined model of human inference and decision-making, enabling a more effective adaptation of the robot's plan.
Lastly, while adaptivity is fundamental to successful collaboration in many types of tasks, it may not lead to improvements in \textit{all} contexts.
For example, in manufacturing scenarios where there is scheduling and allocation ahead of time, adaptation on either side is not required: in these cases, humans actually prefer that robots take the lead and give them instructions \cite{gombolay2014decision}, rather than adapting to their actions.

\textbf{Conclusions.} Overall, our results demonstrate
the benefits of combining motion-level inference with task-level plan adaptation in the context of human-robot collaboration.
Although our results are based on virtual simulations, we speculate that the improvement found here will be even more important in real-world situations, during which humans may be under considerable stress and cognitive load. In such cases, a robot partner not behaving as expected can lead to frustration or confusion, impeding the human's judgment and performance. A robot partner that adapts to the human can make all the difference, both in the successful completion of the task and in the perceived quality of the interaction.

\bibliographystyle{abbrv}
\bibliography{library} 

\begin{thebibliography}{10}

\bibitem{Baker2009}
C.~L. Baker, R.~Saxe, and J.~B. Tenenbaum.
\newblock Action understanding as inverse planning.
\newblock {\em Cognition}, 113(3):329--349, 2009.

\bibitem{bemelmans2012socially}
R.~Bemelmans, G.~J. Gelderblom, P.~Jonker, and L.~De~Witte.
\newblock Socially assistive robots in elderly care: a systematic review into
  effects and effectiveness.
\newblock {\em Journal of the American Medical Directors Association},
  13(2):114--120, 2012.

\bibitem{Brogan2003}
D.~Brogan and N.~Johnson.
\newblock {Realistic human walking paths}.
\newblock {\em IEEE International Workshop on Program Comprehension}, 2003.

\bibitem{charniak1993bayesian}
E.~Charniak and R.~P. Goldman.
\newblock A {B}ayesian model of plan recognition.
\newblock {\em Artificial Intelligence}, 64(1):53--79, 1993.

\bibitem{dias2008sliding}
M.~B. Dias, B.~Kannan, B.~Browning, E.~G. Jones, B.~Argall, M.~F. Dias,
  M.~Zinck, M.~M. Veloso, and A.~J. Stentz.
\newblock Sliding autonomy for peer-to-peer human-robot teams.
\newblock {\em Intelligent Autonomous Systems (IAS)}, 2008.

\bibitem{dragan2015effects}
A.~D. Dragan, S.~Bauman, J.~Forlizzi, and S.~S. Srinivasa.
\newblock Effects of robot motion on human-robot collaboration.
\newblock {\em ACM/IEEE International Conference on Human-Robot Interaction
  (HRI)}, 2015.

\bibitem{dragan2012formalizing}
A.~D. Dragan and S.~S. Srinivasa.
\newblock Formalizing assistive teleoperation.
\newblock {\em Robotics, Science and Systems (RSS)}, 2012.

\bibitem{dragan2013policy}
A.~D. Dragan and S.~S. Srinivasa.
\newblock A policy-blending formalism for shared control.
\newblock {\em The International Journal of Robotics Research}, 32(7):790--805,
  2013.

\bibitem{fern2014decision}
A.~Fern, S.~Natarajan, K.~Judah, and P.~Tadepalli.
\newblock A decision-theoretic model of assistance.
\newblock {\em Journal of Artificial Intelligence Research}, 50(1):71--104,
  2014.

\bibitem{fong2005peer}
T.~Fong, I.~Nourbakhsh, C.~Kunz, L.~Fl{\"u}ckiger, J.~Schreiner, R.~Ambrose,
  R.~Burridge, R.~Simmons, L.~M. Hiatt, A.~Schultz, et~al.
\newblock The peer-to-peer human-robot interaction project.
\newblock {\em AIAA Space}, 2005.

\bibitem{furukawa2006recursive}
T.~Furukawa, F.~Bourgault, B.~Lavis, and H.~F. Durrant-Whyte.
\newblock Recursive {B}ayesian search-and-tracking using coordinated {UAV}s for
  lost targets.
\newblock {\em IEEE International Conference on Robotics and Automation
  (ICRA)}, 2006.

\bibitem{gombolay2014decision}
M.~C. Gombolay, R.~A. Gutierrez, G.~F. Sturla, and J.~A. Shah.
\newblock Decision-making authority, team efficiency and human worker
  satisfaction in mixed human-robot teams.
\newblock {\em Robotics, Science and Systems (RSS)}, 2014.

\bibitem{grocholsky2006cooperative}
B.~Grocholsky, J.~Keller, V.~Kumar, and G.~Pappas.
\newblock Cooperative air and ground surveillance.
\newblock {\em IEEE Robotics \& Automation Magazine}, 13(3):16--25, 2006.

\bibitem{Gureckis15}
T.~M. Gureckis, J.~Martin, J.~McDonnell, R.~S. Alexander, D.~B. Markant,
  A.~Coenen, J.~B. Hamrick, and P.~Chan.
\newblock {psiTurk}: An open-source framework for conducting replicable
  behavioral experiments online.
\newblock {\em Behavioral Research Methods}, 2015.

\bibitem{gurobi}
{Gurobi Optimization}.
\newblock Gurobi reference manual, 2015.
\newblock Available at \url{http://www.gurobi.com.}

\bibitem{held1970traveling}
M.~Held and R.~M. Karp.
\newblock The traveling-salesman problem and minimum spanning trees.
\newblock {\em Operations Research}, 18(6):1138--1162, 1970.

\bibitem{Hoffman2013}
G.~Hoffman.
\newblock Evaluating fluency in human-robot collaboration.
\newblock {\em ACM/IEEE International Conference on Human-Robot Interaction
  (HRI), Workshop on Human Robot Collaboration}, 2013.

\bibitem{hoffman2007effects}
G.~Hoffman and C.~Breazeal.
\newblock Effects of anticipatory action on human-robot teamwork efficiency,
  fluency, and perception of team.
\newblock {\em ACM/IEEE International Conference on Human-Robot Interaction
  (HRI)}, 2007.

\bibitem{kidd2006sociable}
C.~D. Kidd, W.~Taggart, and S.~Turkle.
\newblock A sociable robot to encourage social interaction among the elderly.
\newblock {\em IEEE International Conference on Robotics and Automation
  (ICRA)}, 2006.

\bibitem{lenz2008joint}
C.~Lenz, S.~Nair, M.~Rickert, A.~Knoll, W.~Rosel, J.~Gast, A.~Bannat, and
  F.~Wallhoff.
\newblock Joint-action for humans and industrial robots for assembly tasks.
\newblock {\em Robot and Human Interactive Communication (RO-MAN)}, 2008.

\bibitem{Liu2014a}
C.~Liu, S.-y. Liu, E.~L. Carano, and J.~K. Hedrick.
\newblock {A framework for autonomous vehicles with goal inference and task
  allocation capabilities to support task allocation with human agents}.
\newblock {\em Dynamic Systems and Control Conference (DSCC)}, 2014.

\bibitem{McRuer1995}
D.~T. McRuer.
\newblock {\em Pilot-induced oscillations and human dynamic behavior}, volume
  4683.
\newblock NASA, 1995.

\bibitem{pezzulo2013human}
G.~Pezzulo, F.~Donnarumma, and H.~Dindo.
\newblock Human sensorimotor communication: a theory of signaling in online
  social interactions.
\newblock {\em PLoS ONE}, 8(2), 2013.

\bibitem{Saffarian2012a}
M.~Saffarian, J.~C.~F. de~Winter, and R.~Happee.
\newblock {Automated Driving: Human-Factors Issues and Design Solutions}.
\newblock {\em Proceedings of the Human Factors and Ergonomics Society Annual
  Meeting}, 56(1):2296--2300, 2012.

\bibitem{Sarter1997}
N.~B. Sarter and D.~D. Woods.
\newblock {Team play with a powerful and independent agent: operational
  experiences and automation surprises on the Airbus A-320.}
\newblock {\em Human Factors}, 39(4):553--569, 1997.

\bibitem{shah2010empirical}
J.~Shah and C.~Breazeal.
\newblock An empirical analysis of team coordination behaviors and action
  planning with application to human--robot teaming.
\newblock {\em Human Factors}, 52(2):234--245, 2010.

\bibitem{tisdale2009autonomous}
J.~Tisdale, Z.~Kim, and J.~K. Hedrick.
\newblock Autonomous uav path planning and estimation.
\newblock {\em IEEE Robotics \& Automation Magazine}, 16(2):35--42, 2009.

\bibitem{tomasello2005understanding}
M.~Tomasello, M.~Carpenter, J.~Call, T.~Behne, and H.~Moll.
\newblock Understanding and sharing intentions: the origins of cultural
  cognition.
\newblock {\em Behavioral and Brain Sciences}, 28(05):675--691, 2005.

\bibitem{vesper2010minimal}
C.~Vesper, S.~Butterfill, G.~Knoblich, and N.~Sebanz.
\newblock A minimal architecture for joint action.
\newblock {\em Neural Networks}, 23(8):998--1003, 2010.

\bibitem{wada2007living}
K.~Wada and T.~Shibata.
\newblock Living with seal robots -- its sociopsychological and physiological
  influences on the elderly at a care house.
\newblock {\em IEEE Transactions on Robotics}, 23(5):972--980, 2007.

\bibitem{yen2006agents}
J.~Yen, X.~Fan, S.~Sun, T.~Hanratty, and J.~Dumer.
\newblock Agents with shared mental models for enhancing team decision makings.
\newblock {\em Decision Support Systems}, 41(3):634--653, 2006.

\bibitem{ziebart2009planning}
B.~D. Ziebart, N.~Ratliff, G.~Gallagher, C.~Mertz, K.~Peterson, J.~A. Bagnell,
  M.~Hebert, A.~K. Dey, and S.~Srinivasa.
\newblock Planning-based prediction for pedestrians.
\newblock {\em IEEE/RSJ International Conference on Intelligent Robots and
  Systems (IROS)}, 2009.

\end{thebibliography}
\end{document}